\documentclass{article}
\pdfoutput=1

\PassOptionsToPackage{numbers, compress}{natbib}

\usepackage[preprint]{neurips_2024}

\usepackage{makecell}
\usepackage{multirow}
\usepackage[utf8]{inputenc} 
\usepackage[T1]{fontenc}    
\usepackage{threeparttable}
\usepackage{hyperref}       
\usepackage{url}            
\usepackage{booktabs}       
\usepackage{amsfonts}       
\usepackage{nicefrac}       
\usepackage{microtype}      
\usepackage{subfigure}
\usepackage{soul}
\usepackage[table,xcdraw]{xcolor}
\usepackage{colortbl}

\usepackage{graphicx}
\usepackage{bbm}
\usepackage[linesnumbered,ruled,vlined]{algorithm2e}

\usepackage{caption}
\usepackage{subcaption}
\usepackage{textcomp}
\usepackage{amsmath}

\usepackage{wrapfig}

\newcommand{\model}{\textsc{TCTO}}
\DeclareMathOperator*{\argmax}{argmax} 

\title{Enhancing Tabular Data Optimization with a Flexible Graph-based Reinforced Exploration Strategy}

\author{%
\textbf{Xiaohan Huang}\\
CNIC, CAS\\
University of CAS\\\And
Dongjie Wang\\
University of Kansas\\
\And \textbf{Zhiyuan Ning}\\
CNIC, CAS\\
University of CAS\\\And
\textbf{Ziyue Qiao}\\
Great Bay University\\\And
\textbf{Min Wu}\\
I$^2$R,
A*STAR\\
\And
\textbf{Qingqing Long}\\
CNIC, CAS\\\And
\textbf{Haowei Zhu}\\
Tsinghua University\\
\And
\textbf{Yuanchun Zhou}\\
    CNIC, CAS\\
\And \textbf{Meng Xiao}\thanks{Corresponding Author}\\
    CNIC, CAS\\
    \texttt{shaow@cnic.cn}\\
}

\begin{document}

\maketitle

\begin{abstract}
Tabular data optimization methods aim to automatically find an optimal feature transformation process that generates high-value features and improves the performance of downstream machine learning tasks.
Current frameworks for automated feature transformation rely on iterative sequence generation tasks, optimizing decision strategies through performance feedback from downstream tasks.
However, these approaches fail to effectively utilize historical decision-making experiences and overlook potential relationships among generated features, thus limiting the depth of knowledge extraction.
Moreover, the granularity of the decision-making process lacks dynamic backtracking capabilities for individual features, leading to insufficient adaptability when encountering inefficient pathways, adversely affecting overall robustness and exploration efficiency.
To address the limitations observed in current automatic feature engineering frameworks, we introduce a novel method that utilizes a feature-state transformation graph to effectively preserve the entire feature transformation journey, where each node represents a specific transformation state.
During exploration, three cascading agents iteratively select nodes and idea mathematical operations to generate new transformation states.
This strategy leverages the inherent properties of the graph structure, allowing for the preservation and reuse of valuable transformations.
It also enables backtracking capabilities through graph pruning techniques, which can rectify inefficient transformation paths.
To validate the efficacy and flexibility of our approach, we conducted comprehensive experiments and detailed case studies, demonstrating superior performance in diverse scenarios.
\end{abstract}

\vspace{-0.3cm}
\section{Introduction}
\vspace{-0.3cm}

\begin{figure}[!h]
    \centering
     \vspace{-0.2cm}
\includegraphics[width=0.80\linewidth]{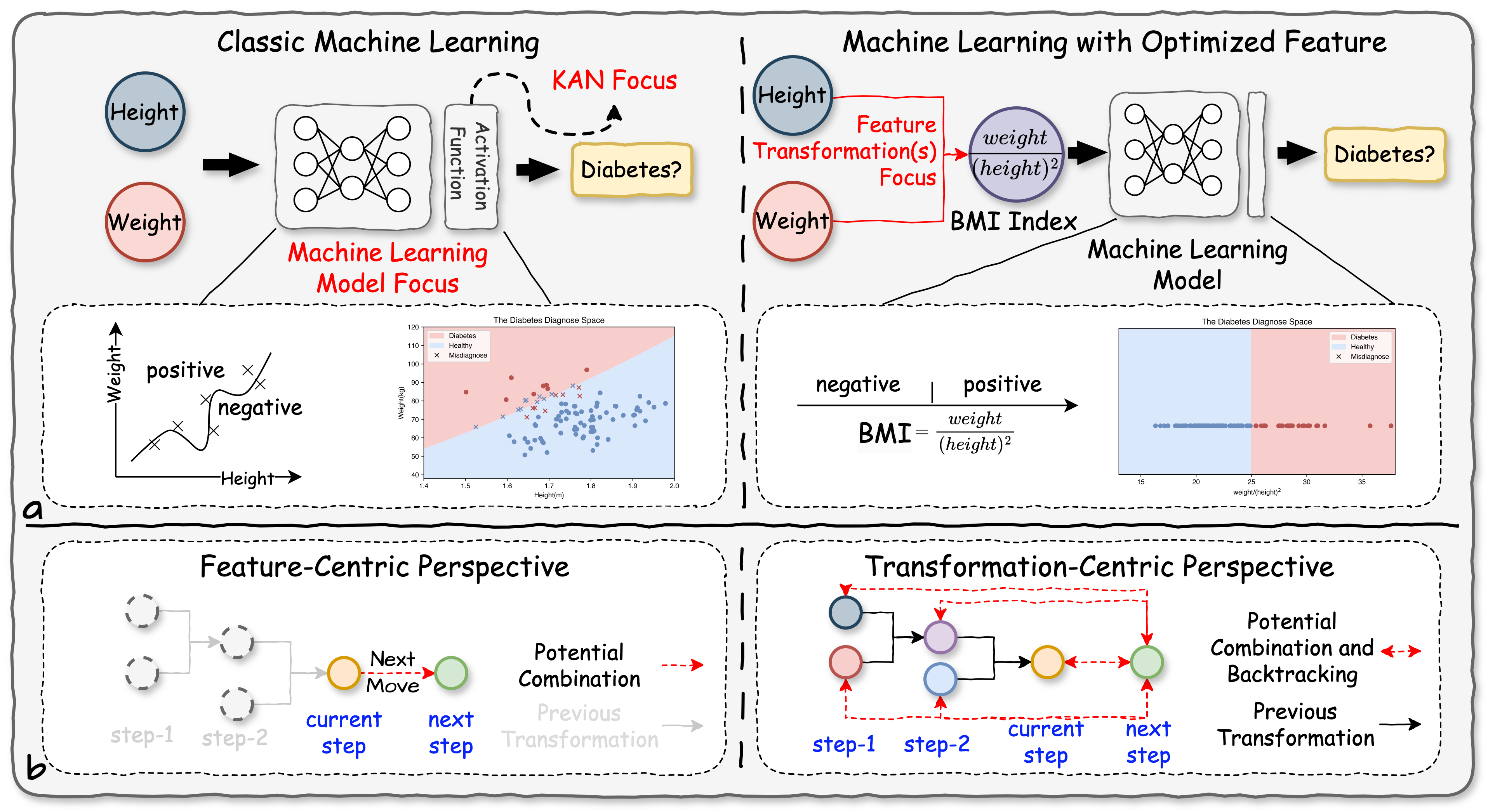}
    \caption{Motivation of this study. (a) Illustration of classic machine learning versus machine learning with optimized features in diabetes diagnosis.
    (b) A conceptual view of feature-centric and transformation-centric perspectives.}
    \label{fig:motivation}
    \vspace{-0.3cm}
\end{figure}

Classic machine learning on tabular data is highly dependent on the structure of the model, the activation function~\cite{liu2024kan}, and most importantly, the quality of the training data~\cite{hancock2020survey, borisov2022deep} (as depicted in Figure~\ref{fig:motivation}(a)). 
Traditionally, optimizing tabular data has required extensive manual intervention by domain experts~\cite{bengio2013representation,conrad2022benchmarking}, which is time-consuming and labor-intensive.
Current research is focused on automatically transforming the original feature spaces through a series of mathematical operations~\cite{zha2023data}, thereby minimizing the reliance on human expertise and streamlining the data preparation phase.  
The mainstream of existing automated feature transformation adopts an iterative perspective: 
\textit{1) expansion-reduction approaches}~\cite{kanter2015deep,khurana2016cognito,horn2019autofeat}  randomly combine and generate features through mathematical transformations, then employ feature selection techniques to isolate high-quality features. 
Those approaches are highly stochastic, lacked stability, and could not learn strategy from transformation steps. 
\textit{2) iterative-feedback approaches}~\cite{tran2016genetic,liu2024interpretable} aim at refining the feature space with the transformation towards reinforcement learning ~\cite{kdd2022,xiao2022traceable,xiao2024traceable} and evolutionary algorithms~\cite{khurana2018feature}. 
Although those methods can optimize and update their strategies during the exploration phase, they discard the valuable experiences from historical sub-transformations and cannot backtrack on individual features. 
\textit{3) AutoML approaches}~\cite{zhu2022difer} partially adjust aforementioned issues by modeling the collected historical transformation records~\cite{wang2024reinforcement} with an autoencoder and then continuously optimize the embedded decision sequence via gradient ascending search.
However, these methods require high-quality search seeds, that is, RL-collected transformation records, which easily result in suboptimal results. 
After thoroughly examining relevant research, a critical question emerges: \textit{How can we develop a framework that effectively reuses high-value sub-transformations, capitalizes on the underlying connections within tabular data, and dynamically adapts its transformation strategy?}

\textbf{Our Perspective and Contributions:} In this work, we pivot from a feature-centric to a transformation-centric approach in addressing the challenges outlined earlier (illustrated in the right section of Figure~\ref{fig:motivation}(b)). This shift brings forth three principal benefits that significantly enhance the capabilities of our reinforcement learning-based automated feature transformation framework:
\textbf{(1) Enhanced Transformation Agility:} Our model is designed to capture and dynamically apply transformations across various stages of the feature transformation process rather than being restricted to transformations derived solely from the current feature set. 
This enables a more flexible and robust handling of features.
\textbf{(2) Historical Insights Utilization:} We leverage deep learning techniques to extract and model latent correlations and mathematical characteristics from past transformation efforts. 
This historical insight informs our decision-making process, allowing the algorithm to organize and execute transformation actions based on the lessons learned strategically.
\textbf{(3) Robust Backtracking Mechanism:} Our approach incorporates a sophisticated backtracking system that utilizes historical transformation records for traceability. 
This feature ensures that the transformation process can revert or alter its course to avoid inefficient or suboptimal trajectories, thus optimizing the overall feature engineering pathway.

\textbf{Summary of Proposed Method: A Transformation-centric Reinforced Tabular Data Optimization Framework.} 
To capitalize on the benefits of a transformation-centric approach, we introduce the \textit{Flexible \textbf{T}ransformation-\textbf{C}entric \textbf{T}abular Data \textbf{O}ptimization Framework} (\textbf{\model}), an innovative automated feature transformation methodology employing a cascading multi-agent reinforcement learning (MARL) algorithm.
Our framework is structured around a dynamic feature-state transformation graph, which is maintained throughout the MARL process.
This graph serves as a comprehensive map, where each node represents a unique sequence of transformations applied to the initial features of the dataset.
Our optimization procedure comprises four steps:
(1) clustering each node on the graph with mathematical and spectral characteristics,
(2) feature transformation-centric cluster state representation,
(3) cluster-level transformation decision generation based on multi-agent reinforcement learning;
(4) evaluation and reward estimation for the generated outcomes.
Iteratively, \model\ executes these steps while leveraging the traceability of the graph for precise node- and stepwise pruning.
This allows for targeted feature reduction and strategic rollbacks, optimizing the transformation pathway. 
Through rigorous experimental validation, we demonstrate that \model\ not only enhances the flexibility of the optimization process but also delivers more resilient and effective results compared to traditional iterative optimization frameworks.


\vspace{-0.3cm}
\section{Preliminary}
\vspace{-0.3cm}
\subsection{Important Definitions}
\vspace{-0.2cm}

\textbf{Tabular Dataset.} 
A tabular dataset is a structured data format that organizes information into rows and columns, similar to a spreadsheet or database table. 
Formally, a tabular dataset can be defined as $\mathcal{D} = [\mathcal{F}, Y]$, where $\mathcal{F}=\{f_1,\dots, f_n\}$ represents $n$ features and $Y$ stands for the labels. 
Each row of $\mathcal{D}$ represents a single observation or data point, while each column corresponds to a specific attribute or feature of the observation. 

\textbf{Operation Set.} 
To enhance the feature space and potentially improve the performance of downstream machine learning models, we can apply a set of mathematical operations to the existing features, generating new and informative-derived features. 
We define this collection of operations as the operation set, represented by the symbol $\mathcal{O}$. 
The operations within this set can be categorized into two main types according to their computational properties: unary and binary operations.
\textit{Unary operations} are those that operate on a single input feature, such as \textit{square}, \textit{exponentiation (exp)}, or \textit{logarithm (log)}. 
\textit{Binary operations} involve two input features and perform operations like \textit{addition}, \textit{multiplication}, or \textit{subtraction}. 

\textbf{Feature-State Transformation Graph.}
A feature-state transformation graph $\mathcal{G}$ is an evolving directed graph and could uniquely represent the dataset optimization process.
This graph structure comprehensively represents the feature transformation process, capturing the relationships between the original features, the intermediate state of the features, and the transformations that generate them.
We can apply the feature-state transformation graph to generate a new dataset $\mathcal{D}'$ with a given tabular dataset, defined as $\mathcal{D}' = \mathcal{G}(\mathcal{D})$.
\begin{wrapfigure}{r}{0.6\textwidth}
  \centering
\includegraphics[width=0.50\textwidth, trim=40 0 30 20]{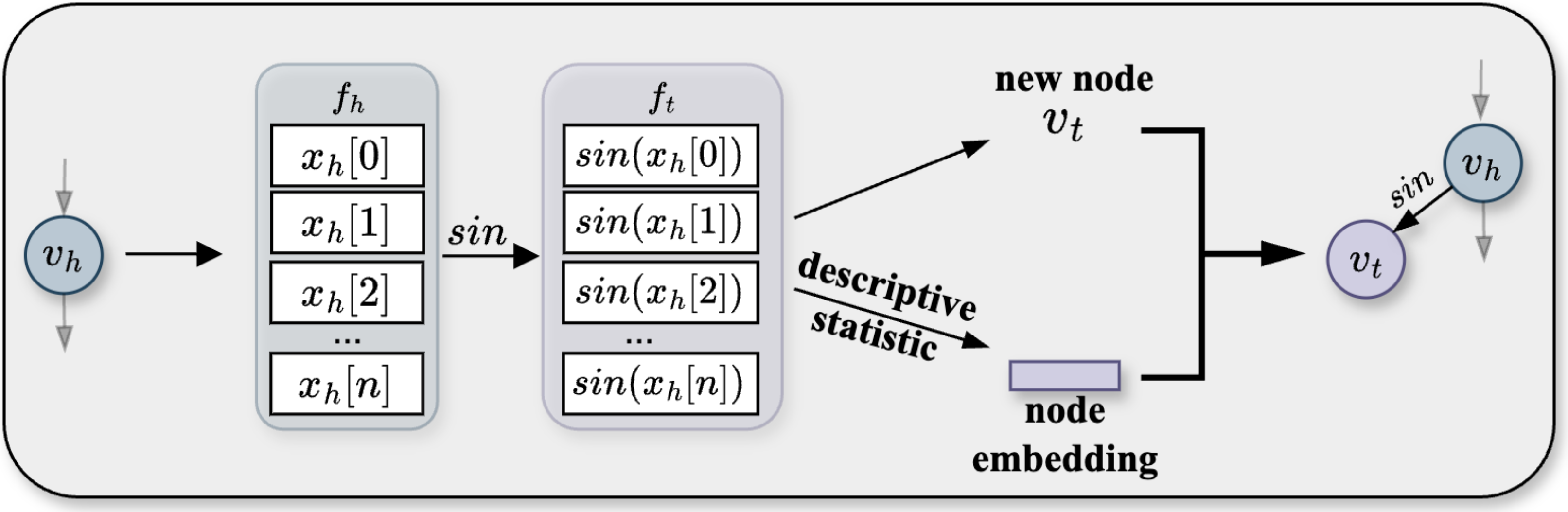}
  \caption{An example of feature-state transformation graph update: the feature $f_h$ conducts $sin$ operation generating the feature $f_t$. 
  The embedding of node $v_t$ can be derived from the statistic description of generated feature $f_t$ .} 
  \label{dem01}
\end{wrapfigure}
This graph $\mathcal{G}  = \{V, E, \mathcal{A}\}$ consists of multiple tree structures where the number of trees equals the number of features in the original dataset. 
$V = \{v_i\}_{i=1}^m$ and $E = \{e_i\}_{i=1}^n$ represent the set of feature state nodes and transformation edges, respectively. $\mathcal{A}$ is the adjacency matrix.
Each pair of nodes, connected by a directed edge, represents a new feature state $v_t$ generated from a previous state $v_h$ after undergoing the transformation represented by the type of edge $e$. 
The embedding of each node will be obtained via the descriptive statistics information (e.g., the standard deviation, minimum, maximum, and the first, second, and third quartile) of the generated features.
Note that in the formulas, \( v \) also represents the embedding of node \( v \) for the sake of simplification.
Figure~\ref{dem01} shows an example of the new generation of edges and nodes. 

\vspace{-0.3cm}
\subsection{Tabular Data Optimization Problem}
\vspace{-0.3cm}
As the toy model illustrated in Figure~\ref{fig:motivation}, given a downstream target ML model $\mathcal{M}$ (e.g., classification model, regression model, etc.) and a tabular dataset $\mathcal{D} = [\mathcal{F}, Y]$, our objective is to find an optimal feature-state transformation graph $\mathcal{G}^*$ that can optimize the dataset through mathematical operation in $\mathcal{O}$.
Formally, the objective function can be defined as:
\begin{equation}
    \label{objective}
    \mathcal{G}^{*} = \argmax_{\mathcal{G}} \mathcal{V}( \mathcal{M}(\mathcal{G}(\mathcal{F})),Y),
\end{equation}
where $\mathcal{V}$ denotes the evaluation metrics according to the target downstream ML model $\mathcal{M}$.

\vspace{-0.3cm}
\section{Proposed Method}
\vspace{-0.2cm}
\begin{figure}[!t]
\centering
\includegraphics[width=\linewidth]{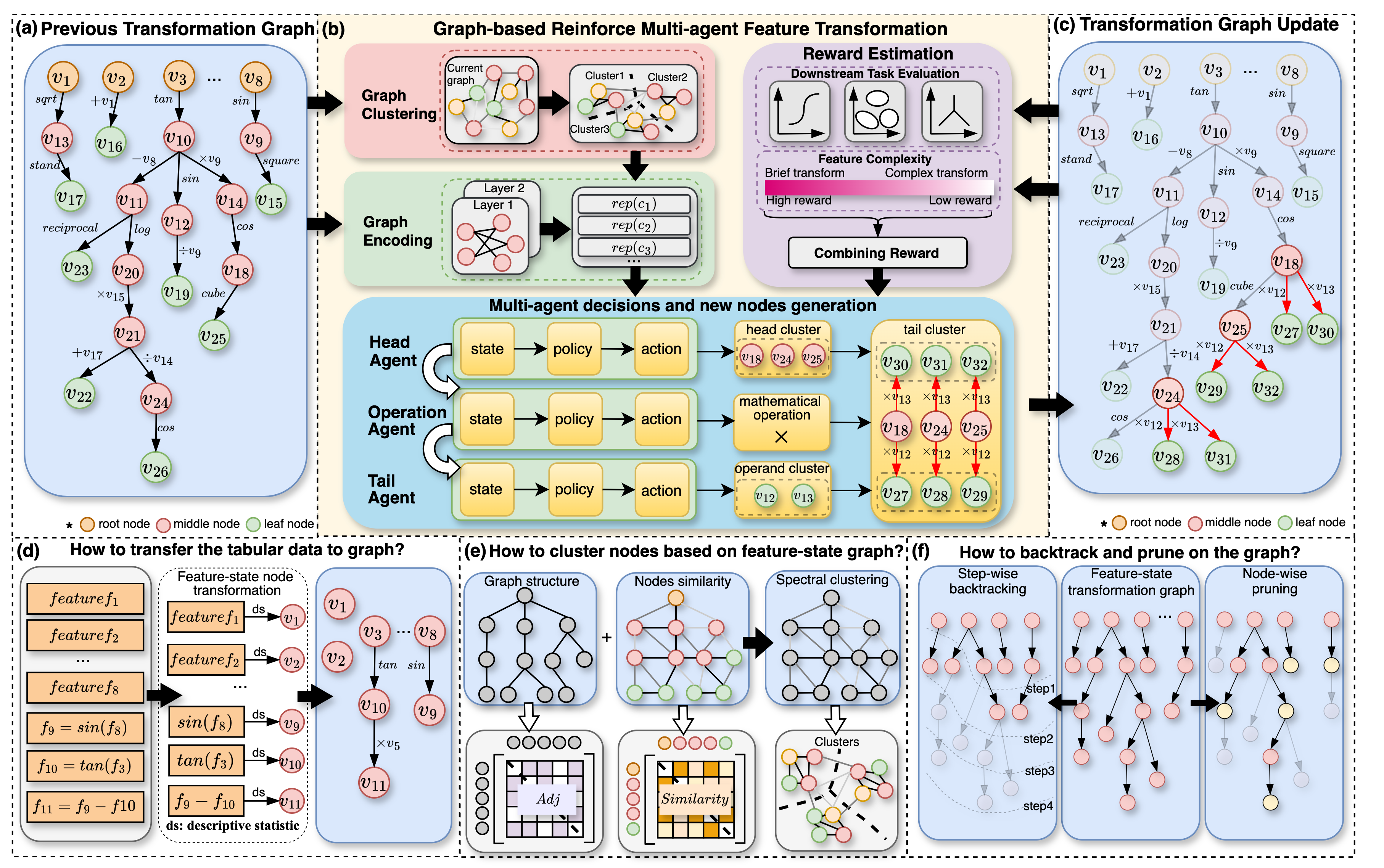}
    \caption{An overview of our framework: (a) construct the feature-state transformation graph based on the previous step; (b) cluster the transformation graph and reinforce multi-agent iterative feature transformation decision generation; (c) update the feature-state transformation graph; (d) details for the process of (transformed) tabular data to feature-state transformation graph; (e) the graph nodes clustering process to form cohesive clusters; (f) illustration of step-wise backtracking and node-wise graph pruning techniques for feature space exploration while maintaining robustness of the pipeline.}
    \vspace{-0.3cm}
    \label{fig:main}
\end{figure}
\subsection{Insights of the Proposed Method}
\vspace{-0.2cm}
Figure~\ref{fig:main} illustrates an overview of our proposed framework, which comprises five key insights:\\
\textbf{1) Effective Transformation Action with Graph Clustering:} The study by GRFG~\cite{kdd2022} shows that the mathematical operation between two distinct groups of features tend to generate high-informative features. 
In contrast, a single feature transformation has little effect on downstream tasks' performance and hinders the optimization of reinforcement learning agents. 
Furthermore, our insight into group-wise operation is that two close features will have similar historical transformation records or mathematical characteristics. 
With the feature-state transformation graph accumulating, this latent relationship will reveal and could be critical to organizing effective yet efficient transformation. 
Implementing graph-based clustering will improve the framework's overall efficacy. \\
\textbf{2) Cluster-level Transformation Decision based on Multi-agent Reinforcement Learning}:
Reinforcement learning has proven effective in solving complex decision-making challenges in numerous domains. 
In our approach, we structure the decision-making process using a cascading system of agents: head, operation, and operand agents. 
These agents operate sequentially to select the optimal head cluster, mathematical operation, and operand cluster, respectively, each according to its learned policy. 
Once the selections are made, the head and operand clusters undergo the specified mathematical operation, creating new nodes within the graph. 
This cascading multi-agent strategy enhances the precision and effectiveness of our transformation decisions, leveraging reinforcement learning's strengths in a novel and impactful way.\\
\textbf{3) Graph-based State Representation for Reinforcement Agent:}  
Achieving an accurate state representation is crucial for enabling reinforcement agents to make informed decisions. 
In our framework, the feature-state transformation graph serves as a repository of extensive intermediate transformation records, complete with their mathematical attributes.
At each step, the agents select clusters of nodes, which can be seen as subgraph components on the feature-state transformation graph.
We then integrate a Relational Graph Convolutional Network (RGCN)~\cite{schlichtkrull2018modeling} to extract and utilize the latent correlations within these historical records and capture the representation of the cluster. 
This approach allows our model to take advantage of the historical insights gained from the RGCN, facilitating strategic transformation actions that are guided by the detailed state of the subgraph. 
This method enhances decision-making precision and significantly improves our algorithm's adaptability in dynamic environments.\\
\textbf{4) Reward Estimation for Optimizing Agents: }
Our model is optimized to generate high-quality features with minimal steps, enhancing efficiency. 
In this context, \model\ evaluates the generated features via the performance of downstream tasks, using this metric as the primary reward to refine the reinforcement learning algorithm. 
Additionally, we factor the complexity of the generated features into the reward calculation. 
This dual focus on performance and complexity ensures that the model aims for effectiveness while maintaining balance, avoiding overly complex solutions that could hinder practical applicability and interoperability.\\
\textbf{5) Effective Graph-Based Backtracking:}
We have implemented two strategic pruning methods to manage the expanding complexity as the number of nodes in our graph grows. 
These approaches are designed to reduce the potential explosion in graph complexity, ensuring the system remains efficient and manageable. 
This backtracking method enhances our system's efficiency and ensures the continual improvement and relevance of the feature transformations.

The following sections will briefly introduce our graph-based reinforced automated feature transformation approach. 
Specifically, we first introduce the graph-related technique components, such as the node clustering, state representation, and graph prune strategy in Section~\ref{graph}. 
Upon that, we illustrate the graph-based cascading reinforcement learning framework, including cascading agents, reward estimation, and optimization in Section~\ref{rl_agent}.
\vspace{-0.2cm}
\subsection{Operation on Dynamic Feature-state Transformation Graph}\label{graph}
\vspace{-0.2cm}
\textbf{Node Clustering on Graph}:
As illustrated in Figure~\ref{fig:main}(e), we delineate the structure of graph $\mathcal{G}$ using its adjacency matrix $\mathcal{A}$, where each element $\mathcal{A}[i, j]$ quantifies the connectivity strength between nodes $v_i$ and $v_j$. Each node in the graph is characterized by an embedding vector that encapsulates its feature information, denoted by the same notation $v$ for simplicity.
To enhance our analysis of inter-node relationships, we compute a similarity matrix $\tilde{\mathcal{A}}$ based on the cosine similarity between the embedding vectors of the nodes. The cosine similarity is calculated as follows:
\begin{equation}
\tilde{\mathcal{A}}[i, j] = \frac{v_i \cdot v_j}{\|v_i\| \|v_j\|}
\end{equation}
This similarity matrix $\tilde{\mathcal{A}}$ is integrated with the adjacency matrix $\mathcal{A}$ to amalgamate structural and feature-based information, thereby augmenting the efficacy of clustering or other graph analytical tasks.
Furthermore, we define an enhanced Laplacian matrix $\mathcal{S}$ to capture both structural and mathematical information from nodes, formulated as follows:
\begin{equation}
\mathcal{S} = \mathcal{D} - (\mathcal{A} +  \tilde{\mathcal{A}})
\end{equation}
Here, $\mathcal{D}$ represents the degree matrix, with diagonal elements $\mathcal{D}[i, i]$ equal to the sum of the elements in the $i$-th row of $\mathcal{A} + \tilde{\mathcal{A}}$.
The clustering module uses hierarchical clustering based on the eigenvalues and eigenvectors of $\mathcal{S}$ to identify the optimal partition of the graph into clusters.
The clustering module treats each eigenvector corresponding to node $v_{i}$ as an initial singleton cluster and iteratively merges pairs of shortest clusters to form progressively larger clusters. 
This process continues until the number of clusters reaches a specified number, set to $k$.
The set of clusters is denoted as $C = \{c_i\}_{i=1}^{k}$.

\textbf{Cluster State Representation with Graph}:
As illustrated in Figure~\ref{fig:main}(b), we construct a dual-layer RGCN framework to disseminate and consolidate information across nodes, utilizing various relationship types to accurately represent the state of each cluster, described as:
\begin{equation}
    v_i^{(l+1)} = \phi \left( \sum_{r=1}^{R} \sum_{j \in N(i)} \frac{1}{c_{i,r}} W_r^{(l)} v_j^{(l)} \right)
\end{equation}
where $v_i^{(l)}$ and $v_i^{(l+1)}$ represents the embedding of the $i$-th node in the feature-state transformation graph at RGCN layer-$l$ and layer-$(l+1)$, respectively.
$N(i)$ denotes the set of neighboring nodes of $v_i$, and the degree normalization factor $c_{i,r}$ scales the influence of neighboring nodes. 
$r$ represents the relationships between nodes, which correspond to different mathematical operations.
The resulting sum is then passed through an activation function $\phi$ to produce the final representation of the node $v_i$.
Based on the aggregated node representation, the representation of the cluster $c_i$ can be obtained by $Rep(c_i) = \frac{1}{|c_i|}\sum_{v\in c_i} v$, 
where $|c_i|$ denotes the number of nodes in cluster $c_i$.

\textbf{Graph Prune Strategy}: 
As illustrated in Figure~\ref{fig:main}(f), we employ two pruning strategies to ensure its stability during the graph transformation process.

\textit{1) Node-wise pruning strategy:} entails the identification of $K$ nodes that show the greatest relevance to labels.
This strategy computes the mutual information, defined as the relevance between each node's corresponding features and labels, as follows:
\begin{equation}
    \mathcal{I}(v,Y) = \sum_{f_i \in v} \sum_{y_i \in Y} p(f_i,y_i) \log \frac{p(f_i,y_i)}{p(f_i)p(y_i)}
\end{equation}
where $f_i$ denotes the element values of node $v$ and $y_i$ is its correlated label. 
$\mathcal{I}(v,Y)$ denoted the mutual information based score.
$p(f)$ represents the marginal probability distribution, while $p(f,y)$ represents the joint probability distribution. 
Finally, the framework will select top-$K$ nodes by the score.
The node-wise pruning strategy removes low-correlation nodes while preserving information as much as possible, ensuring exploration diversity.

\textit{2) Step-wise backtracking strategy:} involves tracing back to the previous optimal feature-state transformation graph before the present episode to prevent deviating onto suboptimal paths. 
This stepwise backtracking ensures that the exploration process remains on the correct trajectory by revisiting and affirming the most effective graph configurations.

\textit{3) When and how to prune the graph:} 
Pruning is recommended when the number of nodes in the graph reaches a set threshold.
The node-wise pruning approach helps preserve diversity while minimizing complexity during the initial stages when agents are unfamiliar with the dataset. 
Once agents have grasped the fundamental policy, the step-wise backtracking strategy assumes leadership to enhance exploration stability.
Combining both approaches, the agent explores a sufficiently large search space and maintains stable exploration in the later stages of training. 
Specifically, we adopt node-wise pruning in each step of the initial 30\% of the exploration period, while the subsequent 70\% is equipped with step-wise backtracking.

\vspace{-0.2cm}
\subsection{Reinforcement Learning Framework on the Evolving Graph}\label{rl_agent}
\vspace{-0.2cm}

\textbf{Cascading Reinforcement Learning Agents}: 
A multi-agent reinforcement learning module is developed to select a head cluster, a mathematical operation, and an operand cluster sequentially.

\textit{1) Head Cluster Agent}:
As described earlier, each node on the feature-state transformation graph has been clustered into $\mathcal{C}$.
The first agent aims to select the head cluster to be transformed according to the current state of each cluster. 
Specifically, the $i$-th cluster state is given as $Rep(c_i)$, and the overall state can be represented as $Rep(V)$.
With the head policy network $\pi_h(\cdot)$, the score of select $c_i$ as the action can be estimated by: $s_i^h = \pi_h(Rep(c_i)\oplus Rep(V))$. 
We use $c_h$ to denote the selected cluster with the highest score.

\textit{2) Operation Agent}:
The operation agent aims to select the mathematical operation to be performed according to the graph and selected head cluster. 
The policy network in the operation agent takes $Rep(c_h)$ and the global graph state as input, then chooses an optimal operation from the operation set $\mathcal{O}$: $o = \pi_o(Rep(c_h)\oplus Rep(V))$. 

\textit{3) Operand Cluster Agent}:
If the operation agent selects a binary operation, the operand cluster agent will choose a tail cluster to perform the transformation. 
Similarly to the head agent, the policy network $\pi_t(\cdot)$ will take the state of the selected head cluster, the operation, the general graph state, and the $i$-th candidate tail cluster as input, given as $s_i^t = \pi_t(Rep(c_h)\oplus Rep(V)\oplus Rep(o)\oplus Rep(c_i))$, where $Rep(o)$ is a one-hot embedding for each operation. 
We use $c_t$ to denote the selected tail cluster with the highest score. \\
These aforementioned stages are referred to as one exploration step. 
Depending on the selected head cluster $c_h$, operation $o$, and optional operand cluster $c_t$, \model\ will cross each feature and then update the feature-state transformation graph (as shown in Figure~\ref{dem01} and the pipeline in Figure~\ref{fig:main} (a-c)).

\textbf{Reward Estimation}:
As illustrated in Figure~\ref{fig:main}(b), we reinforced and encouraged the cascading agents to conduct simple yet effective feature transformations. 
Based on this target, we employ the performance of downstream tasks and the complexity of the transformation graph as rewards to optimize the reinforcement learning framework, denoted as $\mathcal{R}_p$ and $\mathcal{R}_c$, respectively. 

\textit{1) Performance of Downstream Tasks: } As the objective in Equation~\ref{objective}, $\mathcal{R}_p$ is calculated as follows:
\begin{equation}
    \mathcal{R}_p = \mathcal{V}(\mathcal{M}(\mathcal{F}_{t+1}), Y) - \mathcal{V}(\mathcal{M}(\mathcal{F}_{t}), Y),
\end{equation}
where $\mathcal{F}_t$ indicates the feature set at the $t$-th step. 

\textit{2) Complexity of the Transformation: } 
The feature complexity reward $\mathcal{R}_c$ is defined as follows:
\begin{equation}
    \mathcal{R}_c = \frac{1}{n} \sum_{j=1}^n \frac{1}{e^{h(v_j)}},
\end{equation}
where $h(v_j)$ represents the number of levels from the root node to node $v_j$ on $\mathcal{G}$.
The total reward $\mathcal{R}$ is defined as follows: $\mathcal{R} = \mathcal{R}_p + \mathcal{R}_c$. 
In each step, the framework assigns the reward equally to each agent that has action. 

\textbf{Optimization of the Pipeline}: 
In the cascading reinforcement learning setup described, the optimization policy is critical to refine the decision making capabilities of the agents involved: the Head Cluster Agent, Operation Agent, and Operand Cluster Agent. The overarching goal of this policy is to iteratively improve the actions taken by these agents to maximize the cumulative rewards derived from both the performance of downstream tasks and the complexity of transformations in the feature-state graph.

\textit{1) Policy Optimization:} The learning process for each agent is driven by a reward mechanism that quantifies the effectiveness and efficiency of the transformations applied to the feature-state graph. Specifically, the optimization policy is framed within a value-based reinforcement learning approach, leveraging a dual network setup architecture: a prediction network and a target network. 
The prediction network generates action-value (Q-value) predictions that guide the agents' decision-making processes at each step. 
It evaluates the potential reward for each possible action given the current state, facilitating the selection of actions that are anticipated to yield the highest rewards.
The target network serves as a stable benchmark for the prediction network and helps to calculate the expected future rewards. 
Decoupling the Q-value estimation from the target values is crucial to reducing overestimations and ensuring stable learning. 

\textit{2) Loss Function: }
The loss function used for training the prediction network is defined as follows:
\begin{equation}
    \mathcal{L} = \mathcal{Q}^\pi_p(s_t, a_t) - \left(\mathcal{R}_t + \gamma \cdot \max_{a_{t+1}} \mathcal{Q}^\pi_t(s_{t+1}, a_{t+1}) \right),
\end{equation}
where prediction network $\mathcal{Q}^\pi_p(s_t, a_t)$ is the Q-value for the current state-action pair from the policy network $\pi(\cdot)$. $\mathcal{R}_t$ is the immediate reward received after taking action $a_t$ in state $s_t$, and $\gamma$ is the discount factor.
$\max_{a_{t+1}} \mathcal{Q}^\pi_t(s_{t+1}, a_{t+1})$ is the maximum predicted Q value for the next state-action pair as estimated by the target network.

The parameters of the prediction network are updated through gradient descent to minimize this loss, thereby aligning the predicted Q values with the observed rewards plus the discounted future rewards. 
To maintain the stability of the learning process, the parameters of the target network are periodically updated by copying them from the prediction network. 

\vspace{-0.3cm}
\section{Experiments}
\vspace{-0.3cm}
This section reports both quantitative and qualitative experiment results between \model, baselines, and ablation variations. 
To thoroughly analyze the multiple characteristics of our approach, we also analyzed the  running time bottleneck, space scalability, robustness under different machine learning models, and case studies of generated features. 
For details of those experiments, please refer to  Appendix~\ref{appendix_exp}. 
For details of the experiment setting, including the dataset description, evaluation metrics, compared methods, hyperparameter settings, and platform information, please refer to Appendix~\ref{exp_setting}.
\vspace{-0.3cm}
\subsection{Overall Comparison}
\vspace{-0.2cm}
This experiment aims to answer the question: \textit{Can our framework generate high-quality features to improve downstream tasks?}
Table~\ref{main_table} presents the overall comparison between our model and other models in terms of F1-score for classification tasks and 1-RAE for regression tasks. 
We observed that our model outperforms other baseline methods in most datasets. 
The primary reason is that it dynamically captures and applies transformations across various stages of the feature transformation process rather than being restricted to the latest nodes, thereby enhancing flexibility and robustness. 
Compared to expansion-reduction, our technique, along with other iterative-feedback methods, demonstrates a significant advantage in performance. 
The fundamental mechanism is that the reinforcement agent is capable of learning and refining its approach to the process, thereby achieving superior performance compared to random exploration. 
Another observation is that our model performs better than other iterative-feedback approaches, such as NFS, TTG, and GRFG.
An explanation could be that our model identifies and incorporates hidden correlations and mathematical properties, enabling it to develop an improved strategy for feature transformation, drawing on extensive historical knowledge from previous efforts. 
Compared with the AutoML-based approach, DIFER, our technique demonstrates a significant improvement. 
This is primarily because DIFER relies on randomly generated transformations, which are unstable and prone to suboptimal results.
Overall, this experiment demonstrates that \model\ is effective and robust across diverse datasets, underscoring its broad applicability for automated feature transformation tasks.

\begin{table*}[!h]
\centering
\vspace{-0.2cm}
\caption{Overall performance comparison. `C' for binary classification, and `R' for regression. The best results are highlighted in \textbf{bold}. The second-best results are highlighted in \underline{underline}. (\textbf{Higher values indicate better performance.})}
\vspace{-0.1cm}
\label{main_table}
\resizebox{\linewidth}{!}{
\begin{tabular}{ccccccccccccccc}
\toprule
Dataset            & Source   & C/R & Samples & Features & RDG  & ERG & LDA & AFAT   & NFS   & TTG  & GRFG  & DIFER  & \model   \\  \midrule
Higgs Boson        & UCIrvine & C  & 50000   & 28 & 0.695 & 0.702 & 0.513 & 0.697 & 0.691 & 0.699 & \textbf{0.709} & 0.669 & $\textbf{0.709}$ \\ \hline
Amazon Employee    & Kaggle   & C   & 32769   & 9  & 0.932 & 0.934 & 0.916 & 0.930 & 0.932 & 0.933 & \underline{0.935} & 0.929 & $\textbf{0.936}$ \\ \hline
PimaIndian         & UCIrvine & C   & 768     & 8 & 0.760 & 0.761 & 0.638 & 0.765 & 0.749 & 0.745 & \underline{0.823} & 0.760 & $\textbf{0.850}$ \\ \hline
SpectF             & UCIrvine & C   & 267     & 44 & 0.760 & 0.757 & 0.665 & 0.760 & 0.792 & 0.760 & \underline{0.907} & 0.766 & $\textbf{0.950}$\\ \hline
SVMGuide3  & LibSVM & C & 1243 & 21 & 0.787 & \underline{0.826} & 0.652 & 0.795 & 0.792 & 0.798 &  \underline{0.836} & 0.773 & $\textbf{0.841}$\\\hline
German Credit      & UCIrvine & C   & 1001    & 24   & 0.680 & 0.740 & 0.639 & 0.683 & 0.687 & 0.645 & \underline{0.745} & 0.656 & $\textbf{0.768}$\\ \hline
Credit Default     & UCIrvine & C   & 30000   & 25     & 0.805 & 0.803 & 0.743 & 0.804 & 0.801 & 0.798 & \underline{0.807} & 0.796 & $\textbf{0.808}$\\ \hline
Messidor\_features & UCIrvine & C   & 1150    & 19   & 0.624 & 0.669 & 0.475 & 0.665 & 0.638 & 0.655 & \underline{0.718} & 0.660 & $\textbf{0.742}$\\ \hline
Wine Quality Red   & UCIrvine & C   & 999     & 12 & 0.466 & 0.461 & 0.433 & 0.480 & 0.462 & 0.467 & \underline{0.568} & 0.476 & $\textbf{0.579}$\\ \hline
Wine Quality White & UCIrvine & C   & 4900    & 12 & 0.524 & 0.510 & 0.449 & 0.516 & 0.525 & 0.531 & \underline{0.543} & 0.507 & $\textbf{0.559}$\\ \hline
SpamBase           & UCIrvine & C   & 4601    & 57 & 0.906 & 0.917 & 0.889 & 0.912 & 0.925 & 0.919 & \underline{0.928} & 0.912 & $\textbf{0.931}$ \\ \hline
AP-omentum-ovary            & OpenML & C   & 275    & 10936  & 0.832 & 0.814 & 0.658 & 0.830 & 0.832 & 0.758 & \underline{0.868} & 0.833 & $\textbf{0.888}$ \\ \hline
Lymphography       & UCIrvine & C   & 148     & 18  & 0.108 & 0.144 & 0.167 & 0.150 & 0.152 & 0.148 & \underline{0.342} &  0.150  & $\textbf{0.389}$\\ \hline
Ionosphere         & UCIrvine & C   & 351     & 34 & 0.912 & 0.921 & 0.654 & 0.928 & 0.913 & 0.902 & \textbf{0.971} & 0.905  & $\textbf{0.971}$\\ \hline
Housing Boston     & UCIrvine & R   & 506     & 13 & 0.404 & 0.409 & 0.020 & 0.416 & 0.425 & 0.396 & \underline{0.465} & 0.381 & $\textbf{0.495}$\\ \hline
Airfoil            & UCIrvine & R   & 1503    & 5  & 0.519 & 0.519 & 0.220 & 0.521 & 0.519 & 0.500 & 0.538 & \underline{0.558}  &$\textbf{0.622}$ \\ \hline
Openml\_618        & OpenML   & R   & 1000    & 50 & 0.472 & 0.561 & 0.052 & 0.472 & 0.473 & 0.467 & \underline{0.589} & 0.408  &$\textbf{0.600}$ \\ \hline
Openml\_589        & OpenML   & R   & 1000    & 25 & 0.509 & 0.610 & 0.011 & 0.508 & 0.505 & 0.503 & \underline{0.599} & 0.463  &$\textbf{0.606}$\\\hline
Openml\_616        & OpenML   & R   & 500     & 50 & 0.070 & 0.193 & 0.024 & 0.149 & 0.167 &  0.156 & \underline{0.467} & 0.076  &$\textbf{0.499}$\\ \hline
Openml\_607        & OpenML   & R   & 1000    & 50 & 0.521 & 0.555 & 0.107 & 0.516 & 0.519 &  0.522 & \underline{0.640} & 0.476  &$\textbf{0.670}$\\ \hline
Openml\_620        & OpenML   & R   & 1000    & 25 &  0.511 & 0.546 & 0.029 & 0.527 & 0.513 & 0.512 &  \underline{0.626} & 0.442  &$\textbf{0.629}$\\ \hline
Openml\_637        & OpenML   & R   & 500     & 50 & 0.136 & 0.152 & 0.043 & 0.176 & 0.152 & 0.144 & \underline{0.289} & 0.072  & $\textbf{0.355}$\\ \hline
Openml\_586        & OpenML   & R   & 1000    & 25 & 0.568 & 0.624 & 0.110 & 0.543 & 0.544 & 0.544 &  \underline{0.650} & 0.482 &$\textbf{0.689}$\\
  \bottomrule
\end{tabular}}
 \begin{tablenotes}
    \tiny
    \item * We reported F1-score for classification tasks, and 1-RAE for regression tasks.
    \end{tablenotes}
    \vspace{-0.6cm}
\end{table*} 
\begin{figure}[!ht]
    \centering
     \vspace{-0.3cm}
    \includegraphics[width=\linewidth]{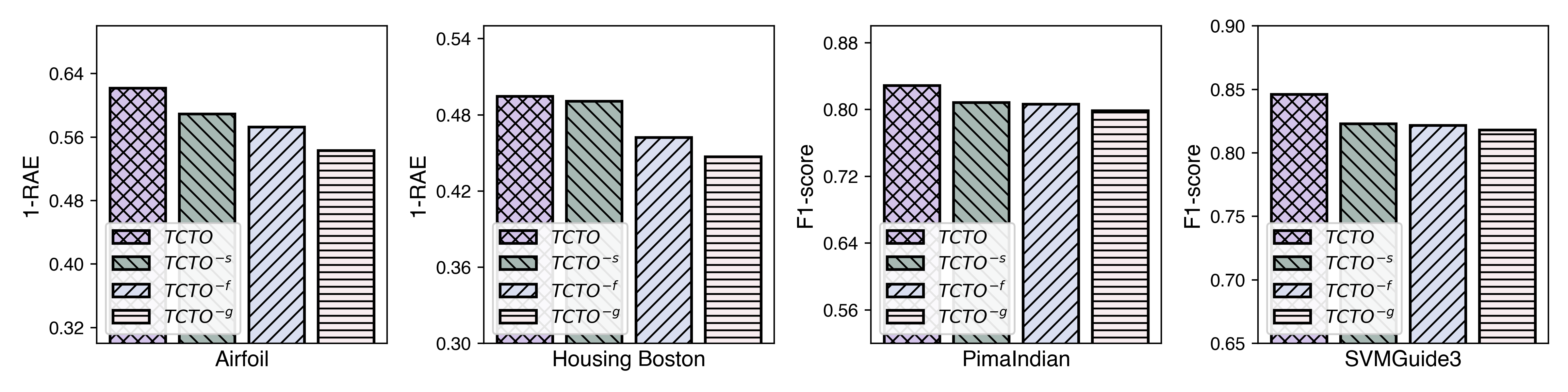}
    \caption{Comparison of TCTO and its variants in Regression and Classification tasks. }
    \vspace{-0.3cm}
    \label{exp:1}
\end{figure}
\begin{figure}[!h]
    \centering
     \vspace{-0.3cm}
\includegraphics[width=\linewidth]{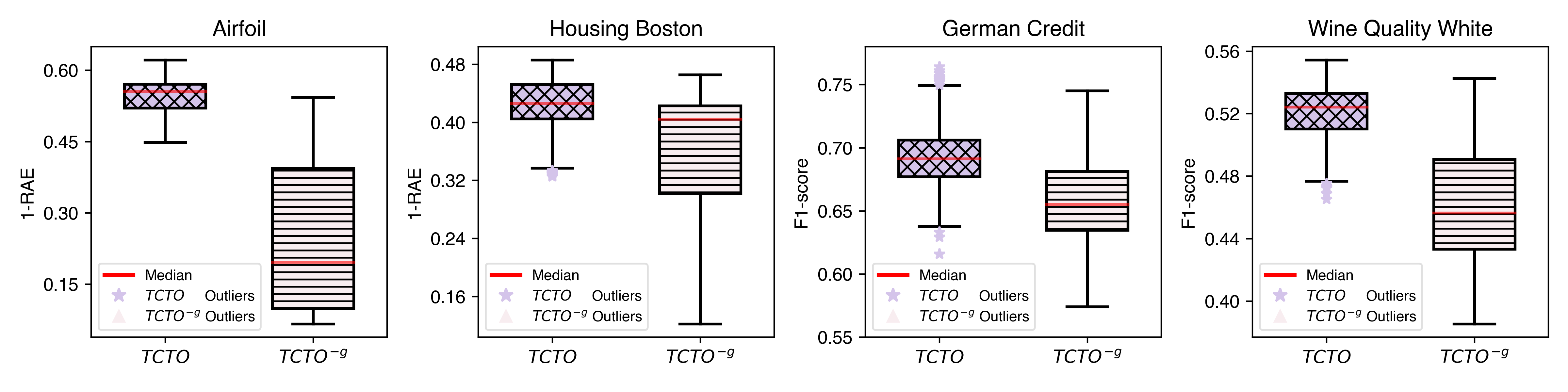}
    \caption{Stability comparison of TCTO and \({\model}^{-g}\) in four different datasets.}
     \vspace{-0.4cm}
    \label{exp:2}
\end{figure} 
\vspace{-0.2cm}
\subsection{Importance of the Feature-state Transformation Graph}
\vspace{-0.2cm}
\label{graphablation}
This experiment aims to answer the question: \textit{How does the feature-state transformation graph impact each component in our model?}
We design three different ablation variants:
1) \textbf{\({\model}^{-f}\)} indicates that the clustering module ignores the mathematical characteristics. 
2) \textbf{\({\model}^{-s}\)} indicates that the clustering module ignores structural information. 
3) \textbf{\({\model}^{-g}\)} ablate the whole graph and then adopt a feature-centric perspective.
The comparison results of these variants are reported in Figure~\ref{exp:1} and Figure~\ref{exp:2}. 

\textbf{Impact on Clustering Component: } 
Figure~\ref{exp:1} illustrates the effectiveness of the optimal features produced by our model and its variants in downstream tasks on the test dataset. 
Firstly, we discovered that \model\ against the other three variants, while ${\model}^{-g}$ showed the weakest performance. 
This indicates that the integration of graph structure and feature information is vital for a percise clustering, which can help the agents to organize transformation between two distinct groups of features, thus generating high-value features. 
We can also observe that ${\model}^{-s}$ outperforms ${\model}^{-f}$ on each task and dataset.
This observation shows that the mathematical characteristic of the generated feature seems to be more significant than structural information. 
The underlying driver is that structural information from historical transformation can enhance the clustering component, thus resulting in better performance (i.e., ${\model}^{-f}$ is superior to ${\model}^{-g}$). 

\textbf{Impact on Cluster State Representation: } 
From Figure~\ref{exp:1}, we can observe a decrease in the performance of downstream tasks when the graph structure is excluded, i.e., \({\model}^{-g}\). 
This performance decline is attributed to the loss of essential information that the feature-state transformation graph maintained. 
In contrast, utilizing the feature-state transformation graph can enable agents to make strategic decisions based on comprehensive historical insights and complex feature interactions. 

\textbf{Impact on Exploration Stability: }\label{stable}
To assess stability, we collected the performance of the downstream task at each exploration step of \model\ and the ablation variation method \({\model}^{-g}\).
Figure~\ref{exp:2} displays box plots summarizing the distributional characteristics of the experimental results.
We can first observe that the median line of our model is consistently higher than ${\model}^{-g}$.
Additionally, the interquartile range (IQR), depicted by the length of the box, indicates that our model's performance distribution is more concentrated than the ablation variation.
The observed stability in our model can be attributed to two primary factors. Firstly, the incorporation of historical and feature information within the graph structure provides guidance, steering the model towards more stable exploration directions. Secondly, the implementation of a graph pruning strategy alongside a backtracking mechanism plays a crucial role; it eliminates ineffective transformed features or reverts the model to the optimal state of the current episode, thereby ensuring stability throughout the exploration process.

\vspace{-0.3cm}
\section{Related Work}
\vspace{-0.3cm}
Feature engineering refers to the process of handling and transforming raw features to better suit the needs of machine learning algorithms \cite{chen2021techniques,zha2023data}. 
Automated feature engineering implies that machines autonomously perform this task without the need for human prior knowledge\cite{lam2017one,zhang2023openfe,cai2023resolving}.
There are three mainstream approaches:
The expansion-reduction based method~\cite{kanter2015deep,horn2019autofeat,khurana2016cognito,lam2017one,khurana2016automating}, characterized by its greedy or random expansion of the feature space\cite{katz2016explorekit,dor2012strengthening}, presents challenges in generating intricate features, consequently leading to a restricted feature space. 
The iterative-feedback approach~\cite{khurana2018feature,tran2016genetic,kdd2022,xiao2022traceable,zhu2022evolutionary,xiao2024traceable} methods integrate feature generation and selection stages into one stage learning process, and aims to learn transformation strategy through evolutionary or reinforcement learning algorithms~\cite{ren2023mafsids}.
However, these methods usually model the feature generation task as a sequence generation problem, ignoring historical and interactive information during the transformation progress, result in lack of stability and flexibility.
The AutoML-based approaches~\cite{wang2021autods,zhu2022difer,xiao2023discrete,ying2023self,ning2024fedgcs} have recently achieved significant advancement.
However, they are limited by the quality of collected transformation and also lack of stability and traceability during the generation phase.
To overcome these problems, \model\ introduces a novel framework that integrates graph-based structural insights and backtracking mechanism with deep reinforcement learning techniques to enhance feature engineering.


\vspace{-0.3cm}
\section{Conclusion Remarks}
\label{limitation}
\vspace{-0.3cm}
In this study, we introduce \model, an automated feature transformation framework. 
Our method emphasizes a transformation-centric approach, in which a feature-state transformation graph is utilized to systematically track and manage feature modifications.
There are three primary advantages of our approach:
(1) Preservation of Transformation Records: The graph structure inherently maintains detailed logs of all feature transformations, which helps to accurately cluster similar features and improves the model's capabilities.
(2) Insightful Decision Making: Using unique structural information and mathematical characteristics, our cascading agents can make informed decisions based on robust state representations.
(3) Robustness through Backtracking: The graph data structure's inherent backtracking capability allows our framework to revert or modify its processing path to avoid inefficient or suboptimal transformation trajectories, thereby enhancing the model’s robustness and adaptability.
Extensive experimental evaluations demonstrate the effectiveness and flexibility of \model\ in optimizing tabular data for a variety of applications.
While \model\ demonstrates significant advancements in automated feature engineering, our analysis has identified a primary bottleneck in the iterative-feedback approach: \textit{the time-consuming nature of downstream task evaluations}. 
This phase often requires extensive computational resources and time, especially when dealing with large datasets and complex models.
In future work, we aim to integrate some unsupervised tabular data evaluation metrics into the \model\ framework, thus making it more suitable for huge datasets.

\newpage
\bibliography{ref}
\bibliographystyle{unsrt}

\newpage
\appendix
\section{Appendix}

\subsection{Supplementary Experiment}\label{appendix_exp}

\subsubsection{Runtime Bottleneck and Temporal Scalability Analysis}\label{time}
\begin{figure}[!h]
\centering 
\subfigure[Time consuming on classification tasks]{
\includegraphics[width=0.45\linewidth]{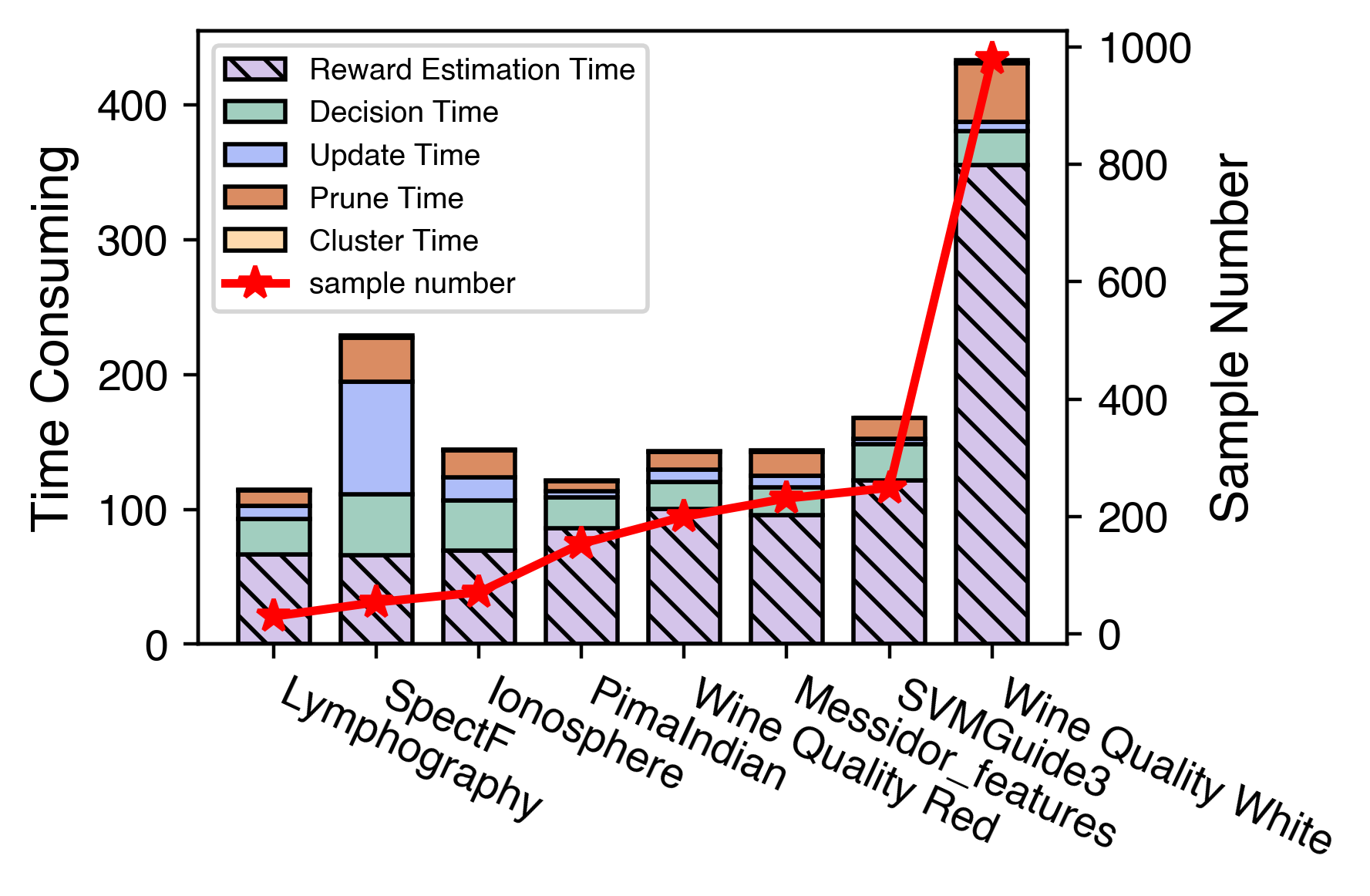}}
\subfigure[Time consuming on regression tasks]{
\includegraphics[width=0.45\linewidth]{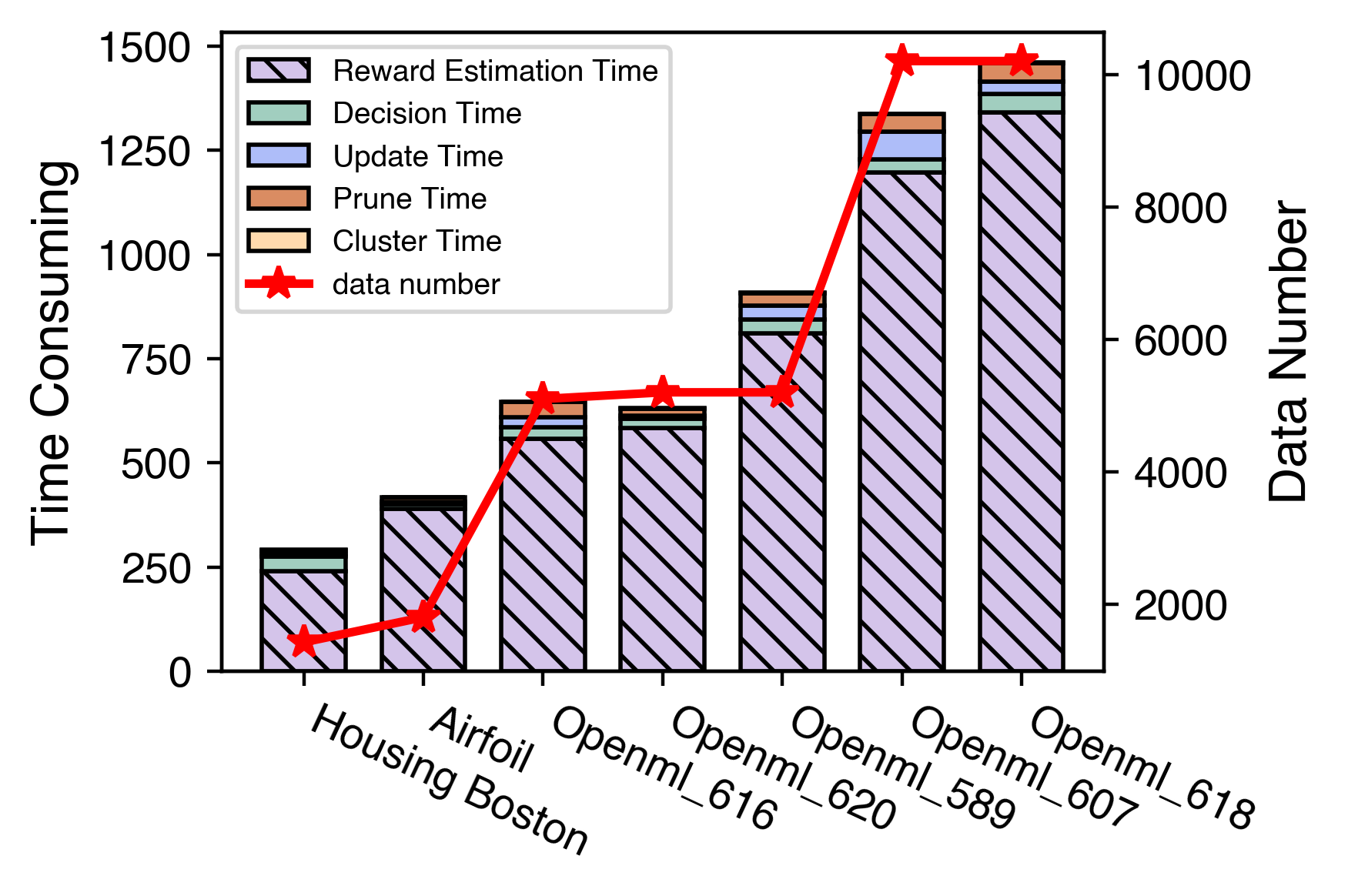}}
\caption{Time consuming of {\model} on different tasks.}
\label{exp:3}
\end{figure}
This experiment aims to answer: \textit{What is the main temporal bottleneck of \model?}
Figure~\ref{exp:3} visualized the average time consumption on each dataset of different modules to analyze the time complexity, including reward estimation, agent decision-making, feature-state transformation graph updating, graph pruning, and graph clustering. 
We can first observe that the reward estimation time dominates the overall time consumption across all dataset sizes. 
This phenomenon can be primarily attributed to the computationally intensive nature of the downstream tasks evaluation process. 
In addition, the time cost of reward estimation increases proportionally with the size of the dataset, which results in a linear scalability of \model\ in terms of time complexity. 
In summary, the main temporal bottleneck of this framework, as well as other iterative-feedback approaches, is the downstream task evaluation in the reward estimation component. 


\subsubsection{Space Complexity Analysis}\label{space}
\begin{table}[!h]
\centering
\caption{The space complexity analysis}
\label{space_complexity}
\resizebox{0.7\linewidth}{!}{
\begin{tabular}{ccccccc}
\toprule
\textbf{\begin{tabular}[c]{@{}c@{}}hidden \\ dimension\end{tabular}} &
  \textbf{\begin{tabular}[c]{@{}c@{}}output \\ dimension\end{tabular}} &
  \textbf{\begin{tabular}[c]{@{}c@{}}embedding \\ dimension\end{tabular}} &
  \textbf{\begin{tabular}[c]{@{}c@{}}head agent \\ parameter\\ size\end{tabular}} &
  \textbf{\begin{tabular}[c]{@{}c@{}}operation agent \\ parameter\\ size\end{tabular}} &
  \textbf{\begin{tabular}[c]{@{}c@{}}operand agent \\ parameter\\ size\end{tabular}} &
  \textbf{\begin{tabular}[c]{@{}c@{}}parameter\\ size\end{tabular}} \\ \midrule
32 & 64 & 7 & 53993 & 14516 & 20213 & 177444 \\
32 & 32 & 7 & 29129 & 8116 & 10613 & 95716 \\
32 & 64 & 16 & 53993 & 14516 & 21257 & 179532 \\
64 & 32 & 7 & 51625 & 8116 & 10613 & 140708 \\
64 & 64 & 7 & 94921 & 14516 & 20213 & 259300 \\
64 & 64 & 16 & 94921 & 14516 & 21257 & 261388 \\ \bottomrule
\end{tabular}}
\end{table}
This experiment aims to answer the question: \textit{Does \model\ have a good spatial scalability?}
Table~\ref{space_complexity} presents the space complexity of each agent and the total number of parameters in {\model}.
Since our model's reinforcement learning structure remains fixed and decoupled with dataset size, it maintains constant space complexity even when exploring large-scale datasets.
We configure various dimensions for RGCN hidden layers, output layers, and operation embeddings to assess their impact on space complexity.
We can observe that the scale of the head cluster agent correlates with the dimensions of RGCN hidden and output layers, as it encodes the feature-state transformation graph during the initial step.
Similarly, the parameter scale of the operation agent is influenced by the dimension of RGCN output layers, as this agent makes decisions based on state information derived from graph embeddings.
The operand cluster agent's space complexity is higher due to its inclusion of an additional embedding layer for encoding mathematical operations within the value network.
Notably, our model employs a dual value-network structure within the deep Q-Learning framework, resulting in a total parameter count twice the sum of the parameters of the three cascading agents.

\subsubsection{Robustness Check}
\begin{table}[htbp]
\centering
\caption{Robustness check of TCTO with distinct ML models on different datasets}
\label{robust}
\begin{minipage}[t]{0.48\linewidth}
\centering
\medskip
\resizebox{\linewidth}{!}{
\begin{tabular}{lccccccc}
\toprule 
      & RFR   & Lasso & XGBR  & SVM-R & Ridge-R & DT-R  & MLP   \\ \midrule
ATF   & 0.433 & 0.277 & 0.347 & 0.276 & 0.187   & 0.161 & 0.197 \\ 
ERG   & 0.412 & 0.162 & 0.331 & 0.278 & 0.256   & 0.257 & 0.300 \\ 
NFS   & 0.434 & 0.169 & 0.391 & 0.324 & 0.261   & 0.293 & 0.306 \\ 
RDG   & 0.434 & 0.193 & 0.299 & 0.287 & 0.218   & 0.257 & 0.279 \\ 
TTG   & 0.424 & 0.163 & 0.370 & 0.329 & 0.261   & 0.294 & 0.308 \\ 
GRFG  & 0.451 & 0.185 & 0.435 & 0.363 & 0.265   & 0.197 & 0.208 \\ 
\textbf{TCTO} & \textbf{0.495} & \textbf{0.370} & \textbf{0.444} & \textbf{0.384} & \textbf{0.317} & \textbf{0.350} & \textbf{0.310} \\ \bottomrule
\end{tabular}}
(a) Housing Boston
\end{minipage}\hfill
\begin{minipage}[t]{0.48\linewidth}
\centering
\medskip
\resizebox{\linewidth}{!}{
\begin{tabular}{lccccccc}
\toprule
     & RFC   & XGBC  & LR    & SVM-C & Ridge-C & DT-C  & KNB   \\ \midrule
ATF  & 0.669 & 0.608 & 0.634 & 0.664 & 0.633   & 0.564 & 0.530 \\
ERG  & 0.683 & 0.703 & 0.659 & 0.571 & 0.654   & 0.580 & 0.537 \\
NFS  & 0.659 & 0.607 & 0.627 & 0.676 & 0.646   & 0.613 & 0.577 \\
RDG  & 0.627 & 0.607 & 0.623 & 0.669 & 0.660   & 0.609 & 0.577 \\
TTG  & 0.650 & 0.607 & 0.633 & 0.676 & 0.646   & 0.599 & 0.577 \\
GRFG & 0.692 & 0.648 & 0.642 & 0.486 & 0.663   & 0.580 & 0.552 \\
\textbf{TCTO} & \textbf{0.742} & \textbf{0.730} & \textbf{0.706} & \textbf{0.701} & \textbf{0.689} & \textbf{0.652} & \textbf{0.587} \\ \bottomrule
\end{tabular}}
(b) Messidor\_features
\end{minipage}
\end{table}
This experiment aims to answer the question: \textit{Are our generative features robust across different machine learning models used in downstream tasks?}
We evaluate the robustness of the generated features on several downstream models.
For regression tasks, we substitute the Random Forest Regressor (RFR) with Lasso, XGBoost Regressor (XGB), SVM Regressor (SVM-R), Ridge Regressor (Ridge-R), Decision Tree Regressor (DT-R), and Multilayer Perceptron (MLP).
For classification tasks, we assess the robustness using Random Forest Classifier (RFC), XGBoost Classifier (XGB), Logistic Regression (LR), SVM Classifier (SVM-C), Ridge Classifier (Ridge-C), Decision Tree Classifier (DT-C), and K-Neighbors Classifier (KNB-C).
Table~\ref{robust} presents the results in terms of 1-RAE for the Housing Boston dataset and F1-score for the Messidor\_features dataset, respectively.
We can observe that the transformed features generated by our model consistently achieved the highest performance in regression and classification tasks among each downstream machine learning method.
Therefore, this experiment validates the effectiveness of our model in generating informative and robust features for various downstream models.

\subsubsection{Case Study on Generated Features}
\begin{table}[!h]
\centering
\caption{Top-10 important features on original and transformed Housing Boston and Wine Quality White datasets}
\label{trace}
\begin{minipage}[t]{\linewidth}
\centering
\medskip
\resizebox{\linewidth}{!}{
\begin{tabular}{rcrccc}
\toprule
\multicolumn{2}{c}{Housing Boston}             & \multicolumn{2}{c}{TCTO$^{-g}$}               & \multicolumn{2}{c}{TCTO}                                   \\
\multicolumn{1}{c}{feature} & importance & \multicolumn{1}{c}{feature} & importance & feature                                       & importance \\ \hline
lstat                       & 0.362      & quan\_trans(lstat)          & 0.144      & \cellcolor[HTML]{C0C0C0}$v_{18}:\sqrt{|v_{17}|}$       & 0.080     \\
rm                          & 0.276      & lstat                       & 0.135      & \cellcolor[HTML]{C0C0C0}$sta(v_{17})$ & 0.077      \\
dis                         & 0.167      & quan\_trans(rm)             & 0.126      & \cellcolor[HTML]{C0C0C0}$sta(\sqrt{|v_{17}}|)$                          & 0.054      \\
crim                        & 0.072      & rm                          & 0.119      & \cellcolor[HTML]{C0C0C0}$sta(v_{16})$                                         & 0.054      \\
rad                         & 0.032      & (dis+(...))-quan(lstat)     & 0.076      & \cellcolor[HTML]{C0C0C0}$sta(\sqrt{\sqrt{v_{18}}})$                   & 0.053      \\
black                       & 0.032      & (dis*(...))+(...)+(dis+...) & 0.050      & \cellcolor[HTML]{C0C0C0}$v_{16}:\frac{1}{\sin{v_{12}}-v_{0}}$                                   & 0.053      \\
age                             & 0.030       & (dis+...)+(...)-(zn+(...))      & 0.048       & $sta(v_{24})$ & 0.050       \\
nox                         & 0.011      & (dis+...)-(...)+quan(rm)    & 0.028      & $\min(v_5)$                                   & 0.044      \\
ptratio                     & 0.007      & (dis+..lstat)-(...+rad)     & 0.016      & \cellcolor[HTML]{C0C0C0}$v_{17}:\sqrt{|v_{16}|}$ & 0.037      \\
indus                       & 0.005      & (dis+..crim)-(...+rad)      & 0.015      & $v_{12}$                    & 0.025      \\ \midrule
\multicolumn{1}{c}{1-RAE:0.414} & Sum:0.993 & \multicolumn{1}{c}{1-RAE:0.474} & Sum:0.757 & 1-RAE:0.494                                                 & Sum:0.527 \\ \bottomrule
\end{tabular}}
\end{minipage}\hfill

\begin{minipage}[t]{\linewidth}
\centering
\medskip
\resizebox{\linewidth}{!}{
\begin{tabular}{rcrccc}
\toprule
\multicolumn{2}{c}{Wine Quality White}             & \multicolumn{2}{c}{TCTO$^{-g}$}                   & \multicolumn{2}{c}{TCTO}                                        \\
\multicolumn{1}{c}{feature} & importance & \multicolumn{1}{c}{feature}     & importance & feature                                            & importance \\ \hline
alcohol                     & 0.118      & quan\_trans(alcohol)            & 0.043      & \cellcolor[HTML]{C0C0C0}$v_{2}+v_{30}$             & 0.026      \\
density                     & 0.104      & alcohol                         & 0.036      & \cellcolor[HTML]{C0C0C0}$\sin{(\sin{(f_{0})})}+v_{30}$ & 0.025      \\
volatile            & 0.099      & ((den...)+(alc...)/(...))       & 0.028      & \cellcolor[HTML]{C0C0C0}$v_{5}+v_{30}$             & 0.024      \\
free sulfur                & 0.093       & quan\_trans(density)               & 0.028       & \cellcolor[HTML]{C0C0C0}$\sin{(f_{0})}+v_{30}$ & 0.023       \\
total sulfur       & 0.092      & density                         & 0.028      & $v_{2}$                                            & 0.023      \\
chlorides                   & 0.091      & (den/(...))+(dens...)/(...)     & 0.026      & \cellcolor[HTML]{C0C0C0}$v_{3}+v_{30}$             & 0.023      \\
residual              & 0.087      & (den/(...)+((...)/tan(...))     & 0.024      & \cellcolor[HTML]{C0C0C0}$v_{6}+v_{30}$             & 0.021      \\
pH                          & 0.082      & (den/...)-(...+stand(...))      & 0.023      & \cellcolor[HTML]{C0C0C0}$v_{7}+v_{30}$             & 0.021      \\
citric acid                 & 0.081      & (citr/(...)+(...)/(tanh(...)) & 0.023      & \cellcolor[HTML]{C0C0C0}$v_{0}+v_{30}$             & 0.021      \\
fixed acidity       & 0.078      & (free/(...)+(...)/tanh(...))   & 0.023      & \cellcolor[HTML]{C0C0C0}$v_{11}+v_{30}$            & 0.021      \\ \midrule
\multicolumn{1}{c}{F1-score:0.536} & Sum:0.924 & \multicolumn{1}{c}{F1-score:0.543} & Sum:0.282 & F1-score:0.559                               & Sum:0.228 \\ \bottomrule
\end{tabular}}
\end{minipage}
\end{table}

This experiment aims to answer the question: \textit{Can our model reuse the high-value sub-transformation and generate a high-quality feature space?}
Table~\ref{trace} presents the Top-10 most important features generated by the original dataset, our proposed method, and its feature-centric variants (i.e., \model$^{-g}$). 
We can first observe that \model\ has reused many high-value sub-transformations, such as node $v_{17}$ in Housing Boston and node $v_{30}$ in Wine Quality White.
Compared to \model$^{-g}$, the graph-based model tends to reuse important intermediate nodes, transforming them to generate more significant features.
A possible reason for this is that our model effectively utilizes historical information from the graph, identifying optimal substructures and exploring and transforming these crucial nodes, thereby utilizing the historical sub-transformations.
Another point to note is that the transformed feature's importance score in our model tends to be more balanced compared to the original dataset and its variant, e.g., the sum of the top-10 feature importance is lower.
Since our model has better performance, we speculate that our framework comprehends the properties of the feature set and ML models to produce numerous significant features by combining the original features. 
Regarding the record of feature transformations shown in Table~\ref{trace}, which is depicted through a formula combining both original and intermediate features, full traceability is also achieved. 
Such characteristics of traceability might help experts find new domain mechanisms.



\subsection{Experiment Settings}\label{exp_setting}

\subsubsection{Experimental Platform Information}
\label{platform}
All experiments were conducted on the Ubuntu 18.04.6 LTS operating system, AMD EPYC 7742 CPU, and 8 NVIDIA A100 GPUs, with the framework of Python 3.8.18 and PyTorch 2.2.0~\cite{pytorch}.

\subsubsection{Dataset and Evaluation Metrics}
\label{datasource}
The datasets utilized for training our model were obtained from publicly accessible repositories, including Kaggle~\cite{kaggle}, LibSVM~\cite{libsvm}, OpenML~\cite{openml}, and the UCI Machine Learning Repository~\cite{uci}. Table~\ref{main_table} provides a succinct summary of these datasets, detailing sample sizes, feature dimensions, and task categories. Our experimental analysis incorporated 14 classification datasets and 9 regression datasets. For evaluation, we utilized the F1-score for classification tasks and the 1-Relative Absolute Error (1-RAE) for regression tasks. In both cases, a higher value of the evaluation metric indicates that the generated features are more discriminative and effective.

\subsubsection{Baseline Methods}
We conducted a comparative evaluation of {\model} against seven other feature generation methods:
(1) \textbf{RDG} randomly selects an operation and applies it to various features to generate new transformed features.
(2) \textbf{ERG} conducts operations on all features simultaneously and selects the most discriminative ones as the generated features.
(3) \textbf{LDA}~\cite{blei2003latent} is a classic method based on matrix decomposition that preserves crucial features while discarding irrelevant ones.
(4) \textbf{AFAT}~\cite{horn2019autofeat} overcomes the limitations of ERG by generating features multiple times and selecting them in multiple steps.
(5) \textbf{NFS}~\cite{chen2019neural} conceptualizes feature transformation as sequence generation and optimizes it using reinforcement learning.
(6) \textbf{TTG}~\cite{khurana2018feature} formulates the transformation process as a graph construction problem at the dataset level to identify optimal transformations.
(7) \textbf{GRFG}~\cite{kdd2022} employs a cascading reinforcement learning structure to select features and operations, ultimately generating new discriminative features.

To ensure experimental integrity, the datasets were divided into training and testing subsets to prevent data leakage. 
The training dataset, comprising 80\% of the data, was used to optimize the reinforcement learning process. 
The testing datasets were used to evaluate the transformation and generation capabilities of the models.
Downstream machine learning tasks were performed using Random Forest Regressor and Random Forest Classifier.

\subsubsection{Hyperparameter Settings and Reproducibility}
\label{hyperparameter}
To comprehensively explore the feature space, we conducted exploration training for 50 episodes, each consisting of 100 steps, during the reinforcement learning agent training phase.
Following training, we assessed the exploration ability of the cascading agents by conducting 10 testing episodes, each comprising 100 steps.
We utilized a two-layer RGCN as the encoder for the feature-transformation state graph, and an embedding layer for the operation encoder.
The hidden state sizes for the graph encoder and operation encoder were set to 32 and 64, respectively.
Each agent was equipped with a two-layer feed-forward network for the predictor, with a hidden size of 100.
The target network was updated every 10 exploration steps by copying parameters from the prediction network.
To train the cascading agents, we set the memory buffer to 16 and the batch size to 8, with a learning rate of 0.01.
For the first 30\% epochs, we employed node-wise pruning strategy to eliminate low-quality features.
Subsequently, we utilized step-wise backtracking strategy for the remaining epochs to restore the optimal feature-state transformation graph.

\end{document}